\documentclass{article}

\usepackage{times}
\usepackage{graphicx} 

\usepackage{natbib}

\usepackage{hyperref}

\usepackage[accepted]{icml2017}

\usepackage{tabulary}
\usepackage{amssymb}
\usepackage{mathtools}
\usepackage{subfig}

\usepackage{tikz}
\usepackage{array}
\usepackage{booktabs}
\usepackage[multiple]{footmisc}
\usepackage{multirow}

\usepackage{xcolor}
\usepackage{algorithm}
\usepackage{algpseudocode}
\usepackage{lscape}

\newcommand{\cut}[1]{}
\newcommand{\bz}{{\bf z}}
\newcommand{\bx}{{\bf x}}
\newcommand{\bm}{{\bf m}}%
\newcommand{\TODO}[1]{\textbf{\textsc{\textcolor{red}{(TODO: #1)}}}}

\icmltitlerunning{Tackling Over-pruning in Variational Autoencoders}

\begin{document} 

\twocolumn[
\icmltitle{Tackling Over-pruning in Variational Autoencoders}



\icmlsetsymbol{equal}{*}

\begin{icmlauthorlist}
\icmlauthor{Serena Yeung}{stanford}
\icmlauthor{Anitha Kannan}{fair}
\icmlauthor{Yann Dauphin}{fair}
\icmlauthor{Li Fei-Fei}{stanford}
\end{icmlauthorlist}

\icmlaffiliation{stanford}{Stanford University, Stanford, CA, USA. Work done during an internship at Facebook AI Research.}
\icmlaffiliation{fair}{Facebook AI Research, Menlo Park, CA, USA}

\icmlcorrespondingauthor{Serena Yeung}{serena@cs.stanford.edu}

\icmlkeywords{unsupervised learning, variational autoencoders}

\vskip 0.3in
]



\printAffiliationsAndNotice{}  

\begin{abstract}
Variational autoencoders (VAE) are directed generative models that learn factorial latent variables. As noted by \citet{BGS16}, these models exhibit the problem of factor over-pruning where a significant number of stochastic factors fail to learn anything and become inactive. This can limit their modeling power and their ability to learn diverse and meaningful latent representations.
In this paper, we evaluate several methods to address this problem and propose a more effective model-based approach called the \emph{epitomic variational autoencoder} (eVAE). The so-called epitomes of this model are groups of mutually exclusive latent factors that compete to explain the data. This approach helps prevent inactive units since each group is pressured to explain the data. We compare the approaches with qualitative and quantitative results on MNIST and TFD datasets. Our results show that eVAE makes efficient use of model capacity and generalizes better than VAE.
\end{abstract}

\section{Introduction}

Unsupervised learning holds the promise of learning the inherent structure in data so as to enable many future tasks including generation, prediction and visualization.  Generative modeling is an approach to unsupervised learning wherein an explicit stochastic generative model of data is defined; independent draws from this model are to produce samples from the underlying data distribution, while the learned latent structure is useful for prediction, classification and visualization tasks. 

Variational autoencoder (VAE) \citep{KW14} is an example of one such generative model.  VAE pairs a top-down generative model with a bottom-up recognition network for amortized probabilistic inference, and jointly trains them to maximize a variational lower bound on the data likelihood.  A number of recent works use VAE as a modeling framework, including  iterative conditional generation of images \citep{GDG+15} and conditional  future frame prediction \citep{XWB+16}.

The generative model of VAE has a set of independent stochastic latent variables that govern data generation; these variables aim to capture various factors of variation. However, a number of studies \citep{BVV+15, SRM+16, KSW+16} have noted that   straightforward implementations  that  optimize the variational bound on the probability of observations converge to a solution in which only a small subset of the stochastic latent units are active.  While it may seem advantageous that the model can  automatically regularize itself, the optimization leads to learning a suboptimal generative model by limiting its capacity to use only a small number of stochastic units. We call this well-known issue with training VAE as 'over-pruning'. Existing methods propose training schemes to tackle the over-pruning problem that arises due to pre-maturely deactivating units \citep{BVV+15, SRM+16, KSW+16}. For instance,  \cite{KSW+16} enforces minimum KL contribution from subsets of latent units while \citep{BVV+15} use KL cost annealing.  However, these schemes are hand-tuned and takes away the principled  regularization scheme that is built into VAE.

We address the over-pruning problem using a model-based approach. We present an extension of VAE called epitomic variational autoencoder (Epitomic VAE, or eVAE, for short) that automatically learns to utilize its model capacity more effectively, leading to better generalization. 
The motivation for eVAE stems from the following observation:   Consider the task of learning a $D$-dimensional representation for the examples in a given dataset. A single example in the dataset can be sufficiently embedded in a smaller $K$-dimensional  $(K\ll D)$ subspace of $D$. However, different data points may need different subspaces, hence the need for $D$. Sparse coding methods also exploit a similar hypothesis. Epitomic VAE exploits sparsity using an additional categorical latent variable in the encoder-decoder architecture of the VAE. Each value of the variable activates only a contiguous subset of latent stochastic variables to generate an observation. This enables learning multiple shared subspaces such that each subspace specializes, and also increases the use of model capacity (Fig.~\ref{fig:activeunits_methods}), enabling better representation. The choice of the name \emph{Epitomic VAE} comes from the fact that multiple miniature models with shared parameters are trained simultaneously.

The rest of the paper is organized as follows. We first describe variational autoencoders and mathematically show the model pruning effect in \S~\ref{sec:vae} and \S~\ref{sec:opvae}. We then present our epitomic VAE model in \S~\ref{sec:approach} that overcomes these shortcomings. Experiments showing qualitative and quantitative results are presented in \S~\ref{sec:exp}. We discuss related work in \S~\ref{sec:related}, and conclude in \S~\ref{sec:conclusion}.

\section{Variational Autoencoders}
\label{sec:vae}
The generative model of a VAE consists of first generating a sample from a D-dimensional stochastic variable $\bz$ that is distributed according to a standard Gaussian:
\begin{equation}
p(\bz) = \prod_{d=1}^{D} \mathcal{N} (z_d; 0, 1)
\end{equation}
Each component $z_i$ captures some latent source of variability in the data. Given $\bz$,  the N-dimensional observation $\bx$ is generated from a parametric family of distributions such as a Gaussian:
\begin{equation}
p_\theta(\bx|\bz) = \mathcal{N} (\bx; f_{1}(\bz), \exp (f_{2}(\bz)))
\end{equation}
where $f_1$ and $f_2$ are non-linear deterministic functions of $\bz$ modeled using neural networks, and $\theta$ denotes the parameters of the generative model. 

The model is trained by optimizing the likelihood $p(X|\theta)$ using a dataset $X$ of $T$ {\it i.i.d.} samples. Since  $p(\bz|\bx)$ is intractable,  VAE approximates the exact posterior using a variational approximation that is amortized across the training set, using a neural network (recognition network) with parameters $\phi$. The resulting variational bound is

\begin{equation}
\footnotesize
\begin{split}
\log &p_{\theta}(X)  = \sum_{t=1}^{T}  \log \int_{\bz} p_\theta(\bx^{(t)},  \bz)\\
& \geq  \sum_{t=1}^{T} E_{q_\phi(\bz|\bx^{(t)})} \log  p(\bx^{(t)}|\bz) - KL\Big[q_\phi(\bz|\bx^{(t)}) \parallel p(\bz)\Big]
\end{split}
\label{eq:bound} 
\end{equation}



 The model is trained using backpropagation to minimize:
\begin{equation}
\footnotesize
\begin{split}
\mathcal{C}_{vae} &=  -\sum_{t=1}^{T} E_{q_\phi(\bz|\bx^{(t)}) } \log  p(\bx^{(t)}|\bz)  \\
& + \sum_{t=1}^{T} \sum_{d=1}^{D} KL \Big(q_\phi(z_d|\bx^{(t)}) \parallel p( z_d)\Big)  
\end{split}
\label{eq:negbound}
\end{equation}
$\mathcal{C}_{vae}$ trade-offs between explaining the data (first term) and ensuring that the posterior distribution is close to the prior $p(\bz)$ (second term). 


\section{Over-pruning}
\label{sec:opvae}
We can better understand over-pruning in the VAE by considering different ways to minimize the sum of two terms in $\mathcal{C}_{vae}$. The first term encourages proper reconstruction while the second term captures the divergence between the posterior $q(z_d)$ and its Gaussian distributed prior $p(z_d)$, independently for each component. The easiest way to minimize the sum is to have a large number of components collapse to the prior $p(z_d)$ to compensate for a few highly non-Gaussian components that help reconstruction. This is achieved by turning off the corresponding component\footnote{log variance is modeled using the neural network, so turning it off to 0 corresponds to a variance of 1.}. This behavior is noticeable in the  early iterations of training when the model for  $\log p(\bx|\bz)$ is  quite impoverished, and improvement to the loss can be easily obtained by optimizing this KL term. However, once the units have become inactive, it is almost impossible to resurrect them.

\begin{figure}
	\centering
	\includegraphics[width=8.0cm]{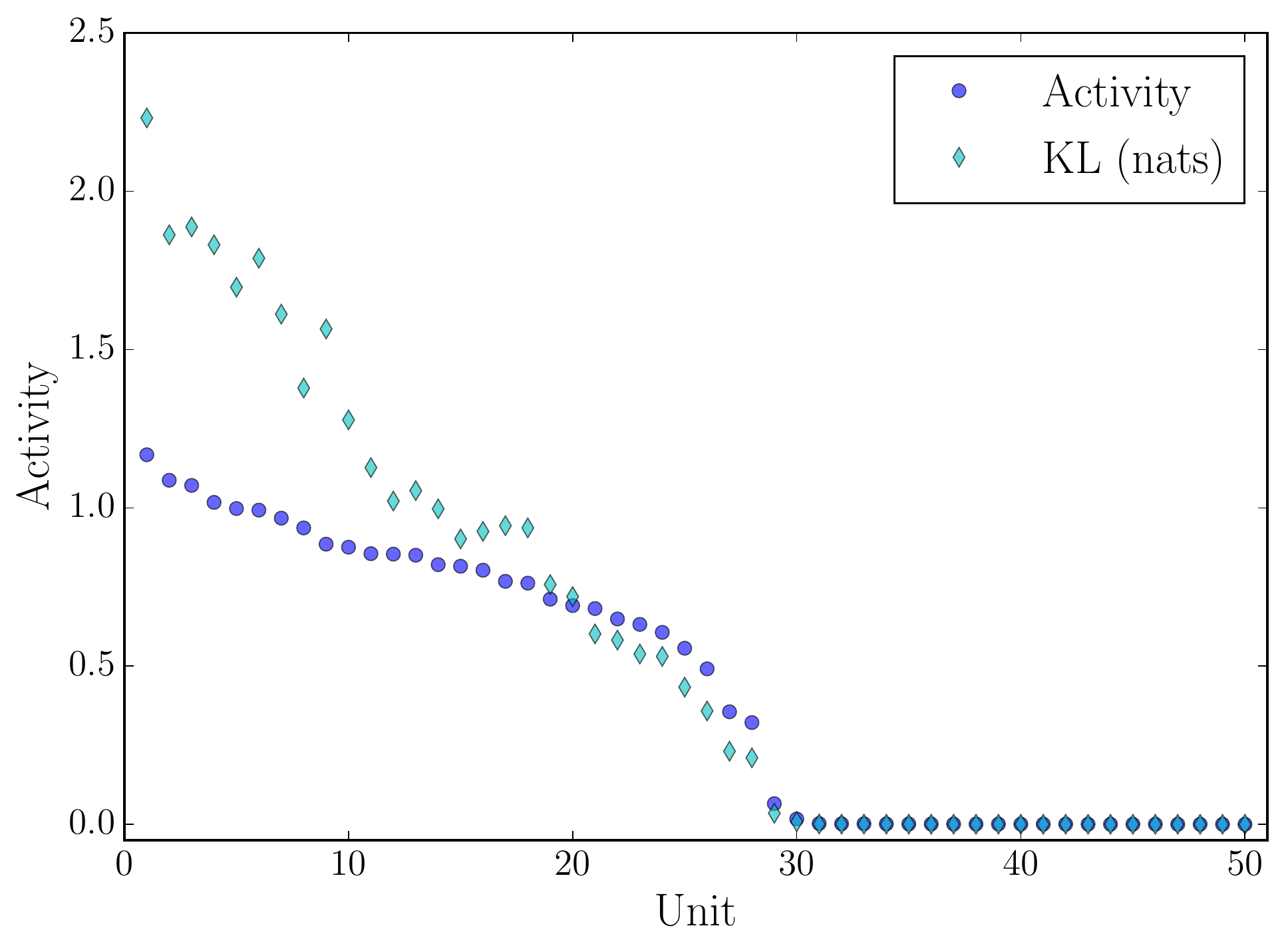}
     \caption{\small{Unit activity and KL term for a 50-unit VAE (sorted by activity). Activity shows correlation with KL term. Highly active units have a high KL with the prior; this mismatch creates a discrepancy between reconstruction and generation performance.}}
  \label{fig:KL_activity}   
\end{figure}

A quantity that is useful in understanding this effect is the activity level of a unit.   Following \cite{BGS16}, we define a unit to be used, or ``active", if $A_u=\mathrm{Cov}_x(\mathbb{E}_{u \sim q(u|\mathbf{x})}[u]) > 0.02$. In Figure~\ref{fig:KL_activity} we plot the activity level and KL-divergence in $\mathcal{C}_{vae}$ for each component of a 50-unit VAE trained on MNIST. This result illustrates that all inactive units have collapsed to the prior, whereas active units are relatively far from the prior.

One could argue that over-pruning is a feature since the model seems to discard unnecessary capacity. However, we observe over-pruning even when the model underfits the data. Moreover, over-pruning creates a discrepancy between training and generation time since it allows some components $q(z_d)$ to be highly different from the prior. Instead, it is desirable for each individual component to be close to the prior, since generation occurs by sampling from the prior.




\label{sec:vae_prune}
\begin{figure}
    \begin{minipage}[c]{8.0cm}
        \subfloat{\includegraphics[width=8.0cm]{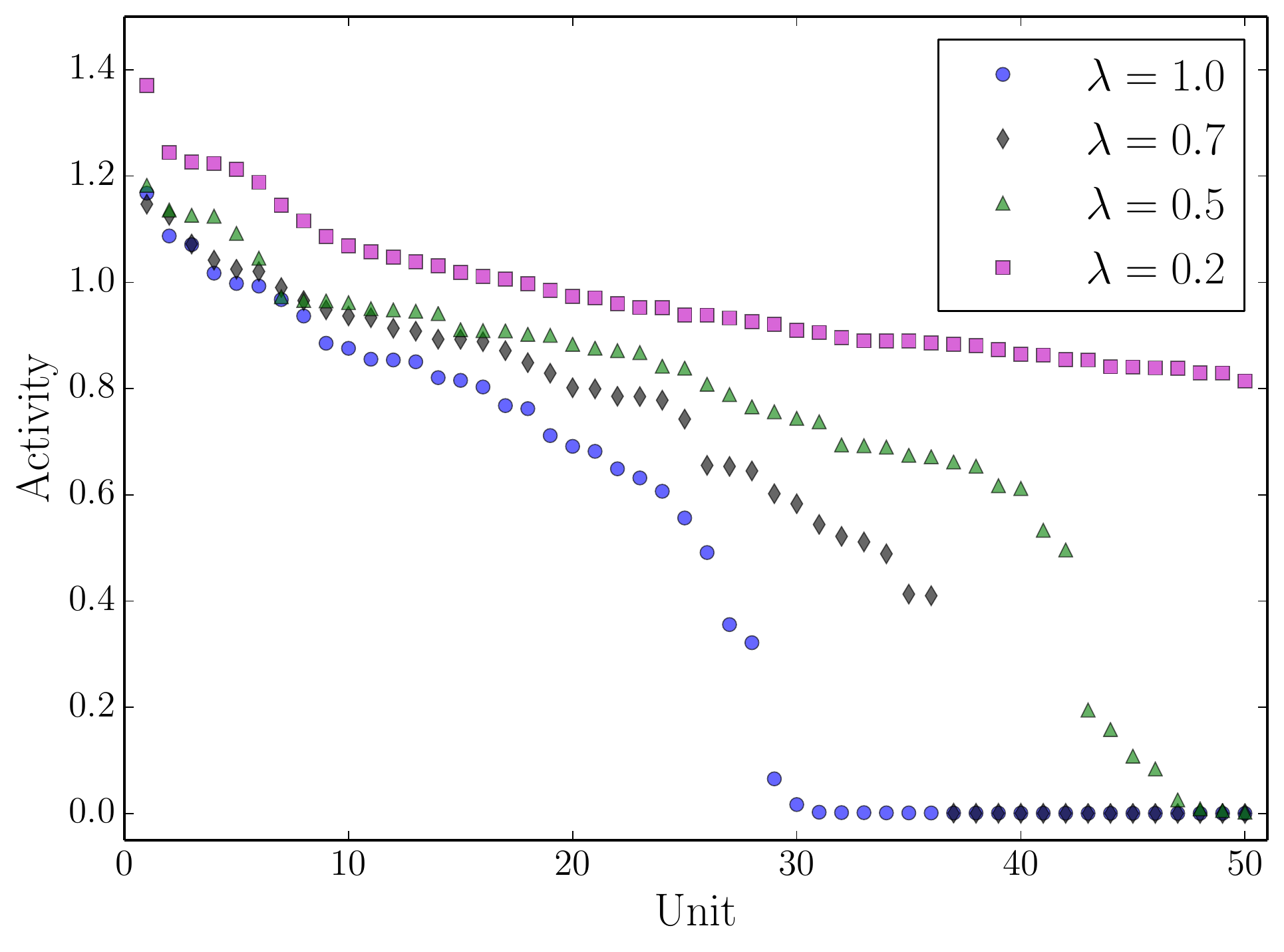}}
    \end{minipage}
    \begin{minipage}[c]{8.2cm}
        \hspace{2pt}
        \subfloat[$\lambda = 1$]{\includegraphics[width=2.5cm]{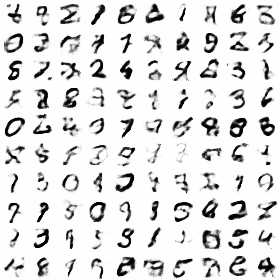}}
        \hspace{5pt}
        \subfloat[$\lambda = 0.5$]{\includegraphics[width=2.5cm]{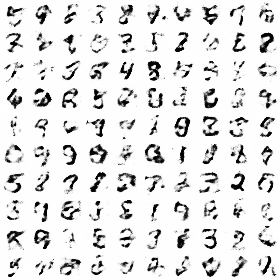}}
        \hspace{5pt}
        \subfloat[$\lambda = 0.2$]{\includegraphics[width=2.5cm]{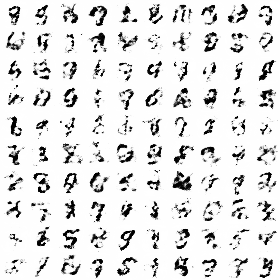}} \\
        \vspace{15pt}
    \end{minipage}
    \caption{\small{Sorted activity of latent units and corresponding generations on MNIST, for a 50-d VAE with a hidden layer of 500 units. Shown for varying values of the KL weight $\lambda$. When $\lambda = 1$, only 30 units are active. As $\lambda$ is decreased, more units are active; however generation does not improve since the model uses the capacity to model increasingly well only regions of the posterior manifold near training samples (see reconstructions in Fig.~\ref{fig:reconstruction_lambda}).}}
    \label{fig:kld_lambda}
\end{figure}

\subsection{Weighting the KL term}
One approach to reducing over-pruning is to introduce a trade-off between the two terms using a parameter $\lambda$:
\begin{equation}
\footnotesize
-E_{q_\phi(\bz|\bx) }[ \log  p(\bx|  \bz) ] +\lambda\sum_{i=1}^{D} KL \Big(q_\phi(z_i|\bx) \parallel p( z_i)\Big)\nonumber
\end{equation}
$\lambda$ controls the importance of keeping the information encoded in $\bz$ close to the prior. $\lambda=0$ corresponds to a vanilla autoencoder, and $\lambda=1$ to the correct VAE objective. Fig.~\ref{fig:kld_lambda} shows the effect of $\lambda$ on unit activity and generation.  While tuning down $\lambda$ increases the number of active units, samples generated from the model are still poor. This is because at small values of $\lambda$, the model becomes closer to a vanilla autoencoder and hence spends its capacity in ensuring that reconstruction of the training set is optimized (sharper reconstructions as a function of $\lambda$ are shown in Appendix \S~\ref{sec:appendix-lambda}), at the cost of generation capability. 


\subsection{Dropout VAE}
\begin{figure}
	\centering
	\includegraphics[width=8.0cm]{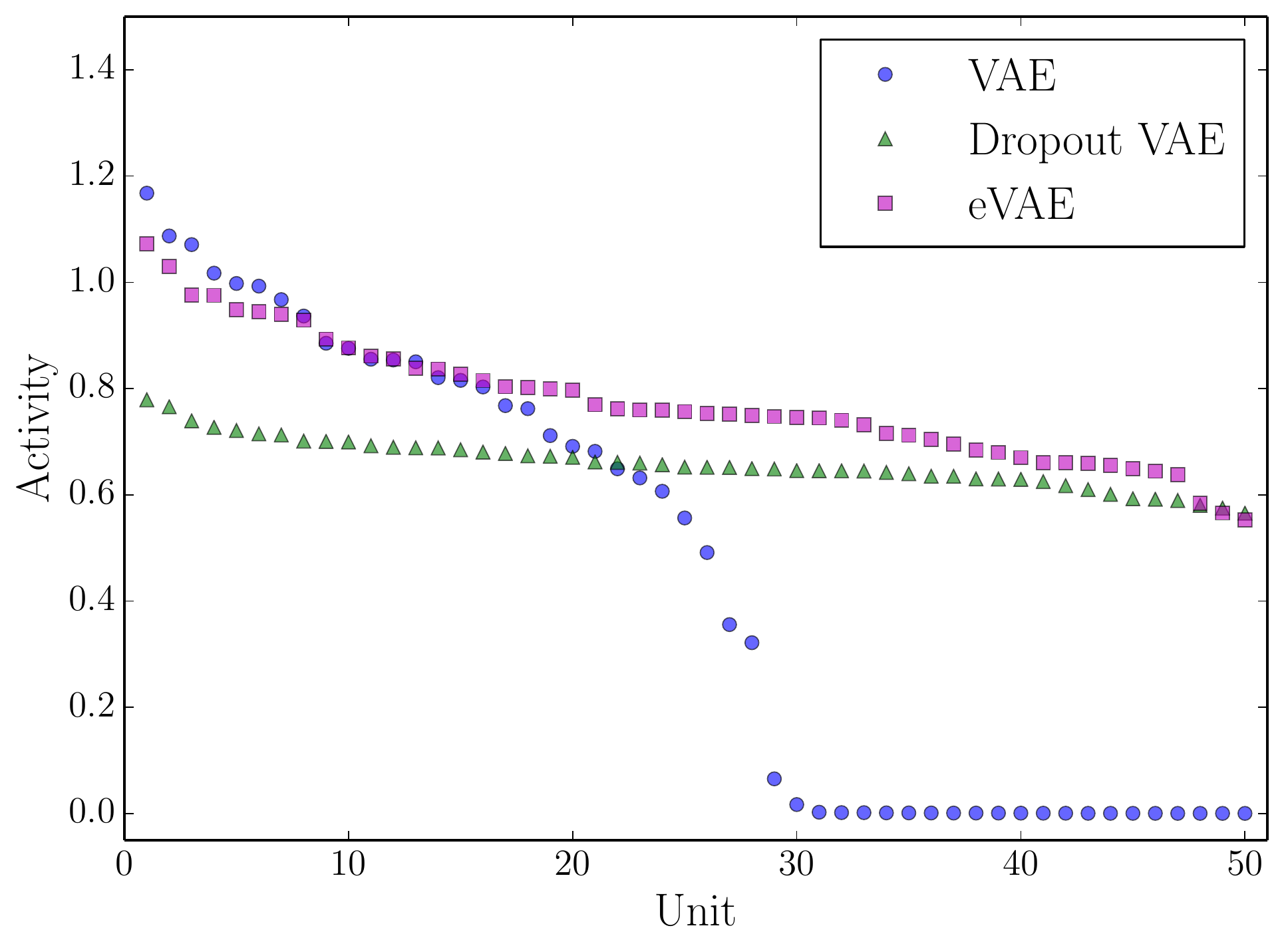}
     \caption{\small{Adding dropout to a VAE (here, dropout rate 0.5 is shown) can prevent the model from pruning units, shown for MNIST. However, in contrast to eVAE, it uses the additional units to encode redundancy, not additional information, and therefore does not address the problem. eVAE is able to utilize the full latent capacity with all units active. Compare generation results for dropout VAE with eVAE in Fig.~\ref{fig:dropout_rec_gen} and Fig.~\ref{fig:mnist_generations}, respectively.}}
  \label{fig:activeunits_methods}   
\end{figure}

\begin{figure}[h]
    \begin{minipage}[c]{8.2cm}
        \hspace{2pt}
        \subfloat[Reconstruction]{\includegraphics[width=3.9cm]{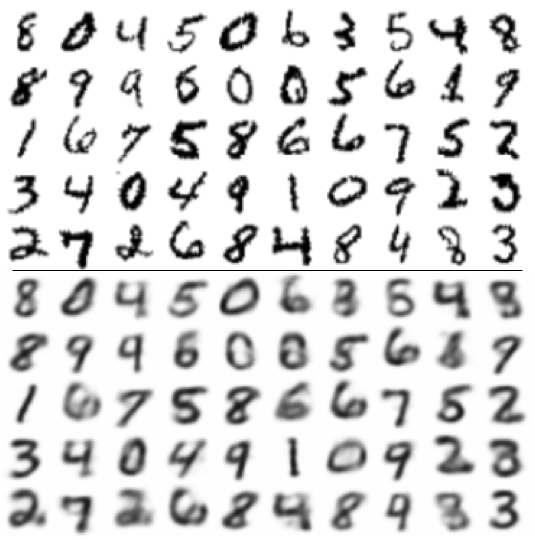}}
        \hspace{5pt}
        \subfloat[Generation]{\includegraphics[width=3.9cm]{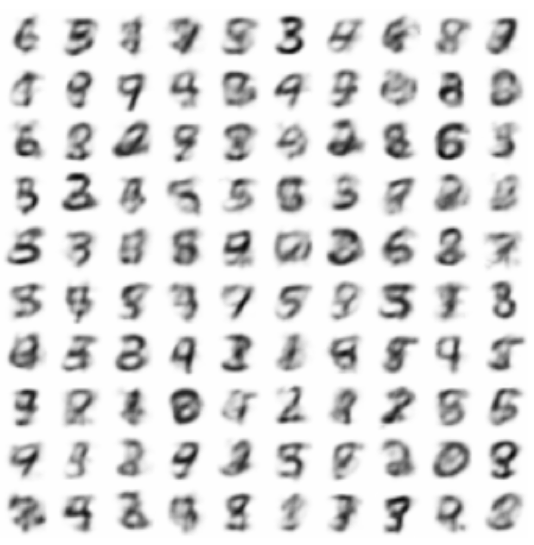}} \\
        \vspace{15pt}
    \end{minipage}
    \caption{\small{Dropout VAE on MNIST: Both generation and reconstruction are  blurry. This is because the additional active units are used to encode redundant information.}}
    \label{fig:dropout_rec_gen}
\end{figure}

Another approach is to add dropout to the latent variable $\bz$ of the VAE (Dropout VAE). While this increases the number of active units (Fig.~\ref{fig:activeunits_methods}), it generalizes poorly as it uses the dropout layers to merely replicate representation.  This results in blurriness in both generation and reconstruction, and illustrates that simply utilizing additional units is not sufficient for proper utilization of these units to model additional factors of variation, as seen in Fig.~\ref{fig:dropout_rec_gen}. 

\section{eVAE: A model-based approach}
\label{sec:approach}

\begin{figure*}
  \centering
  \begin{tabular}{@{}c@{\hskip 0.2in}c@{}}
  \includegraphics[width=0.8\linewidth]{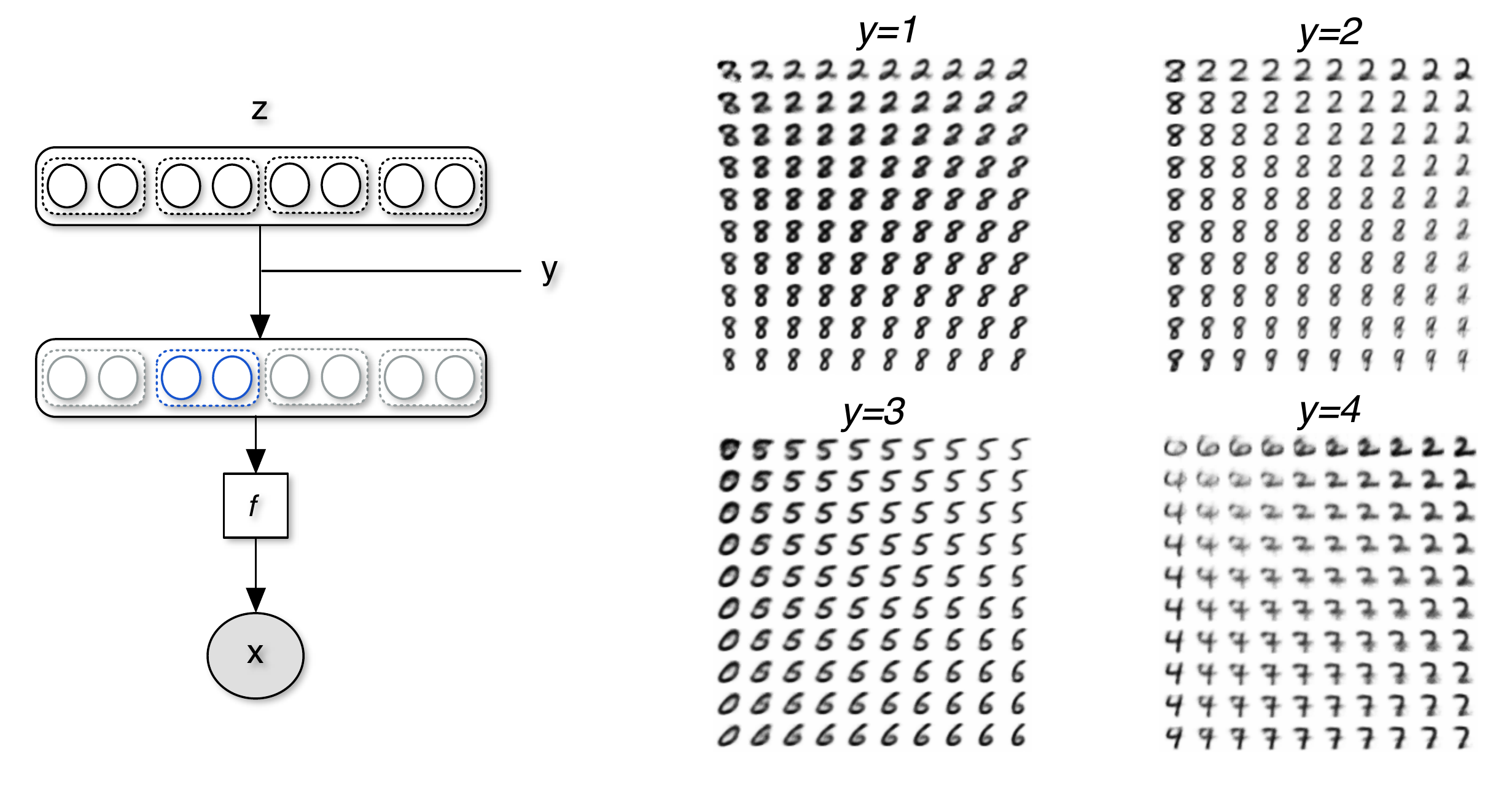}
  \end{tabular}
  \caption{\small{Left: Illustration of an epitomic VAE with dimension D=8, epitome size K=2 and stride S=2. In this depiction, the second epitome is active. Right: Learned manifolds on MNIST for 4 different epitomes in a 20-d eVAE with size $K=2$ and stride $s=1$. We observe that each epitome specializes on a coherent subset of examples; this enables increasing the diversity of the samples generated while maintaining quality of the samples when the latent dimension is large.}}
  \label{fig:model}
\end{figure*}

We propose epitomic variational autoencoders (eVAE) to overcome the over-pruning problem of VAEs. We base this on the observation that while we may need a $D$-dimensional representation to accurately represent every example in a dataset, each individual example can be represented with a smaller $K$-dimensional subspace. As an example, consider MNIST with its  variability in terms of digits, strokes and thickness of ink, to name a few. While the overall $D$ is large, it is likely that only a few $K$ dimensions of $D$ are needed to capture the variability in strokes of some digits (see Fig.~\ref{fig:model}).
Epitomic VAE can be viewed as a variational autoencoder with latent stochastic dimension $D$ that is composed of a number of smaller variational autoencoders  called \emph{epitomes}, such that each epitome partially shares its encoder-decoder architecture with other epitomes in the composition. In this paper, we assume simple structured sparsity for each epitome: in particular, only $K$ \emph{contiguous} dimensions of $D$ are active\footnote{The model also allows for incorporating other forms of structured sparsity.}. 

The generative process can be described as follows:
A  D-dimensional  stochastic variable $\bz$  is drawn from a standard multivariate Gaussian $p(\bz) =\mathcal{N} (\bz; 0, I)$. In tandem, an epitome is implicitly chosen through an epitome selector variable $y$, which has a uniform prior over possible epitomes. The $N$-dimensional observation $\bx$ is then drawn from a Gaussian distribution: 
\begin{equation}
p_\theta(\bx|y,\bz) = \mathcal{N}(\bx; f_1(\bm_y \odot \bz),  \exp(f_2(\bm_y \odot \bz))) \label{eqn:gen}
\end{equation}
$\bm_y$ enforces the epitome constraint:  it is also a $D$-dimensional vector that is zero everywhere except $K$ contiguous dimensions that correspond to the epitome dictated by $y$. $\odot$ is element-wise multiplication between the two operands. Thus, $\bm_y$  masks the  dimensions of $\bz$ other than those dictated by the choice of $y$.   Fig.~\ref{fig:model} illustrates this for an 8-d $\bz$ with epitome size $K=2$, such that there are four possible epitomes (the model also allows for overlapping epitomes, but this is not shown for illustration purposes). Epitome structure is defined using size $K$ and stride $s$, where $s=1$ corresponds to full overlap in $D$ dimensions\footnote{The strided epitome structure allows for learning $O(D)$ specialized subspaces, that when sampled during generation can each produce good samples.  In contrast, if only a simple sparsity prior is introduced over arbitrary subsets (e.g. with Bernoulli latent units to specify if a unit is active for a particular example),  it can lead to poor generation results, which we confirmed empirically but do not report. The reason for this is as follows:  due to an exponential number of potential combinations of latent units, sampling a subset from the prior during generation cannot be straightforwardly guaranteed to be a good configuration for a subconcept in the data, and often leads to uninterpretable samples.}. Our model generalizes the VAE and collapses to a VAE when $D=K=s$. 

$f_1(\diamond)$ and $f_2(\diamond)$ define non-linear deterministic transformations of $\diamond$ modeled using neural networks.  Note that the model does not snip off the $K$ dimensions corresponding to an epitome, but instead  ignores the $D-K$ dimensions that are not part of the chosen epitome.  While the same  deterministic functions $f_1$ and $f_2$ are used for any choice of epitome, the functions can still specialize due to the sparsity of their inputs.  Neighboring epitomes will have more overlap than non-overlapping ones, which manifests itself in the representation space; an intrinsic ordering in the variability is learned. 



\subsection{Overcoming over-pruning}
Following \cite{KW14}, we use a recognition network $q(\bz,y|\bx)$ for approximate posterior inference, with the functional form
\begin{equation}
\begin{split}
q( \bz,y|\bx) &= q(y|\bx)q( \bz|y,\bx) \\
&=  q(y|\bx)\mathcal{N}(\bz; \bm_y \odot \bf{\mu},  \exp{(\bm_y \odot \bf{\phi})})
\end{split}
\label{eq:posterior}
\end{equation}
where $\bf{\mu} = h_1(\bx)$ and $\bf{\phi} = h_2(\bx)$ are  neural networks that map $\bx$ to   $D$-dimensional space.
We use a similar masking operation as  the generative model. Unlike the generative model (eq.~\ref{eqn:gen}), the masking operation defined by $y$ operates directly on outputs of the recognition network that characterizes the parameters of  $q(\bz|y,\bx)$. Similar to VAE,  the lower bound on the log probability of a dataset can be derived, leading to the cost function (negative bound):

\begin{equation}
\footnotesize
\begin{split}
\mathcal{C}_{evae}&= -\sum_{t=1}^{T} E_{q(\bz,y|\bx^{(t)}) }[ \log  p(\bx^{(t)}| y, \bz) ] \\ 
& +\sum_{t=1}^{T} KL  \Big[q_\phi(y|\bx^{(t)}) \parallel p_\theta( y)\Big] \\
& +\sum_{t=1}^{T} \sum_{y} q_\phi(y|\bx^{(t)})  KL \Big[   q_\phi(\bz|y,\bx^{(t)}) \parallel p_\theta( \bz) \Big]
\end{split}
\label{eq:evaebound}
\end{equation}
eVAE  departs from VAE in how the contribution from the KL term is constrained.  Consider the third term expanded:

\begin{equation}
\footnotesize
\begin{split}
& \sum_{t=1}^{T} \sum_{y} q_\phi(y|\bx^{(t)}) KL \Big[   q_\phi(\bz|y,\bx^{(t)}) \parallel p_\theta( \bz) \Big ] \\ 
& = \sum_{t=1}^{T} \sum_{y} q_\phi(y|\bx^{(t)}) \sum_{d=1}^D \mathbf{1}[{m_{d,y} = 1}] KL \Big[q(z_d|\bx^{(t)}) \parallel p(z_d) \Big],
\end{split}
\end{equation}
where $\mathbf{1} [\star] $ is an indicator variable that evaluates to 1 if only if its operand $\star$ is true.
Unlike in VAE where this KL term decomposes into independent KL for each $z_i$, contiguous dimensions of $\bz$ are constrained by the choice of $y$. In addition, the number of KL terms that will contribute to $\mathcal{C}_{evae}$ for an input $\bx^{(t)}$ is exactly $K$ by model design, with the other $D-K$ dimensions set to provide contribution of zero. Thus, only a fraction of examples in the training set contributes a possible non-zero value to $z_d$'s KL term in $\mathcal{C}_{evae}$.  This gives eVAE the ability to use more total units without having to prematurely prune the model to optimize the bound. In contrast, for $\mathcal{C}_{vae}$ to have a small contribution from the KL term of a particular $z_d$, it has to infer that unit to have zero mean and unit variance for  many examples in the training set. In practice, this results in VAE completely inactivating units.

Fig.~\ref{fig:activeunits_methods} compares the activity levels of eVAE with VAE and dropout VAE. Even though  Dropout VAE has similar activity profile to eVAE, its generative model is impoverished as it focuses on replicating representation as opposed to modeling variability. This can be evidenced  by comparing the generation results from the two models,  in Fig.~\ref{fig:dropout_rec_gen} and Fig.~\ref{fig:mnist_generations}.


\subsection{Training}
The generative model and the recognition network are trained simultaneously, by minimizing $\mathcal{C}_{evae}$ in Eq.~\ref{eq:evaebound}. 

For the stochastic continuous variable $\bz$, we use the reparameterization trick as in VAE, which reparameterizes the recognition distribution in terms of auxiliary variables with fixed distributions. This allows efficient sampling from the posterior distribution as it becomes a deterministic function of inputs and auxiliary variables. 

For the discrete variable $y$, we cannot use the reparameterization trick. We therefore approximate $q(y|\bx)$ by a point estimate $y^*$ so that  $q(y|\bx)=\delta(y-y^*)$, where $\delta$ evaluates to $1$ only if $y=y^*$ and the best $y^* = \arg \min \mathcal{C}_{evae}$. We also explored modeling $q(y|\bx) = Mult(h(\bx))$  as a discrete distribution with $h$ being a neural network.  In this case, the backward pass requires either using REINFORCE or passing through gradients for the categorical sampler. In our experiments, we found that these approaches did not work well, especially when the number of possible values of $y$ becomes large. We leave this as future work to explore. 

The recognition network first computes  $\bf{\mu}$ and $\bf{\phi}$. It is then combined with the optimal $y^*$ for each example, to arrive at the final posterior.  The  model is trained using a simple algorithm outlined in Alg.~\ref{alg:evae}. Backpropagation with minibatch updates is used, with each minibatch constructed to be balanced with respect to epitome assignment.

\begin{algorithm}
\caption{Learning Epitomic VAE}
\label{alg:evae}
\begin{algorithmic}[1]
\State $\theta$, $\phi$ $\leftarrow$Initialize parameters
\For {until convergence of parameters  ($\theta$, $\phi$) }
\State Assign each $\bx$ to its best $y^* = \arg \min \mathcal{C}_{evae}$ 
\State Randomize and partition data into minibatches, with each minibatch having proportionate number of examples $\forall$ $y$ 
  \For{ k $\in$  numbatches }
    \State Update model parameters using $k^{th}$ minibatch consisting of  $\bx, y$ pairs 
 \EndFor
\EndFor
\end{algorithmic}

\end{algorithm}

\section{Experiments}
\label{sec:exp}
We present experimental results on two datasets, MNIST~\citep{lecun1998mnist} and Toronto Faces Database (TFD)~\citep{susskind2010toronto}. We use standard splits for both MNIST and TFD. In our experiments, the encoder and decoder are fully-connected networks, and we show results for different depths and number of units of per layer. ReLU nonlinearities are used, and models are trained using the Adam update rule~\citep{kingma2014adam} for 200 epochs (MNIST) and 250 epochs (TFD), with base learning rate 0.001. We emphasize that in all experiments, we optimize the correct lower bound for the corresponding models. 


\subsection{Qualitative results: Reconstruction vs. Generation}
\label{sec:exp-overpruning}

\begin{figure*}[t]
	\centering
    \includegraphics[width=0.98\linewidth]{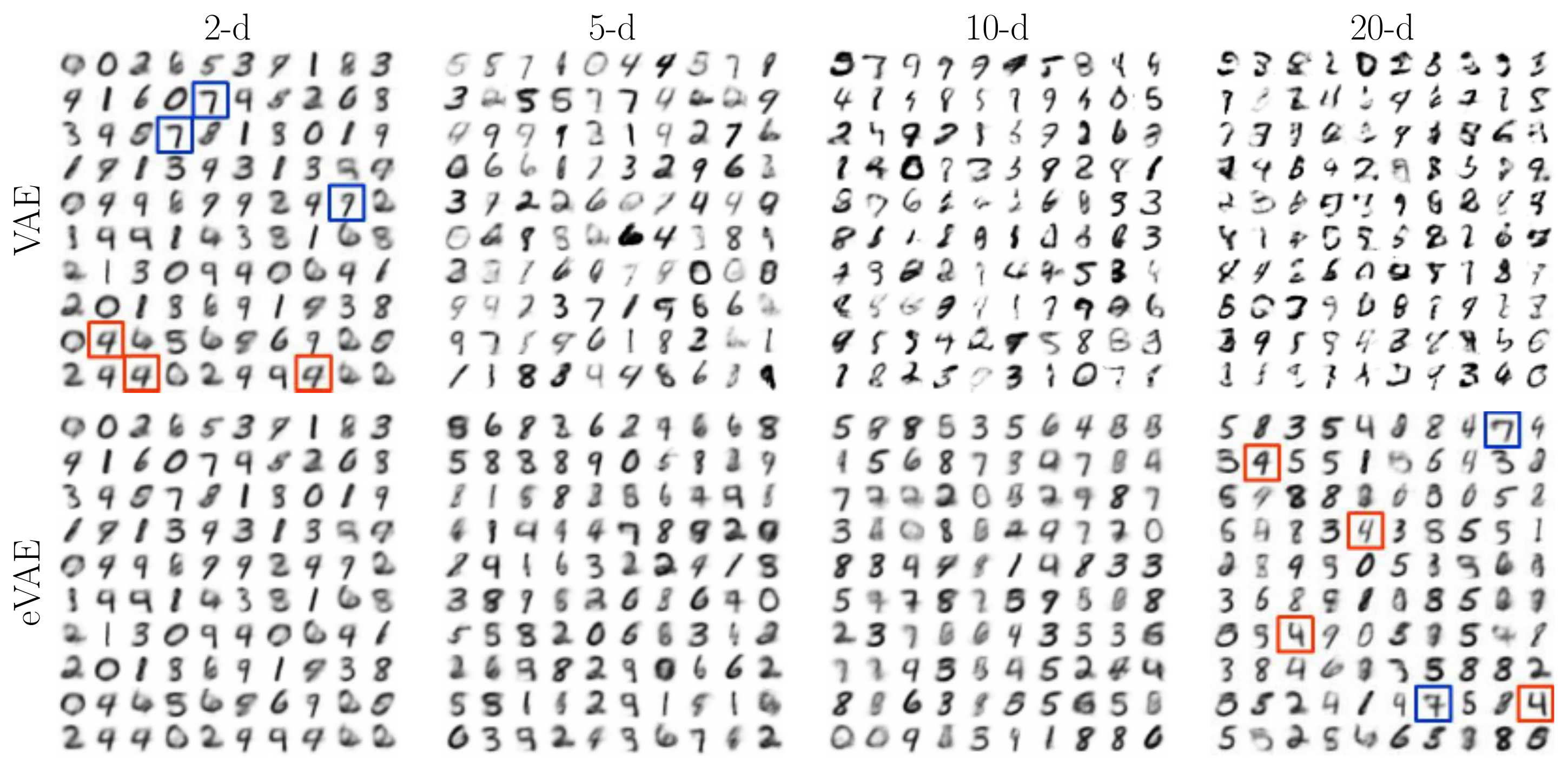}
    \caption{\small Generations from VAE and eVAE models for different dimensions of latent variable $\bz$. In this experiment, we maintain a simple encoder-decoder architecture with a single layer of 500 deterministic units (samples from best architecture are in Fig.~\ref{fig:sample_generations}).  Across each row are 2-d, 5-d, 10-d, and 20-d models. VAE generation quality degrades as latent dimension increases, and it is unable to effectively use added capacity to model greater variability. eVAE overcomes the problem by modeling multiple shared subspaces, here 2-d (overlapping) epitomes are maintained as the latent dimension is increased. Learned epitome manifolds from the 20-d model are shown in Fig.~\ref{fig:model}. Boxed digits highlight the difference in variability that the VAE vs. eVAE model is able to achieve.}
  	\label{fig:mnist_generations}
\end{figure*}

\begin{figure}[ht]
	\centering
    \includegraphics[width=\linewidth]{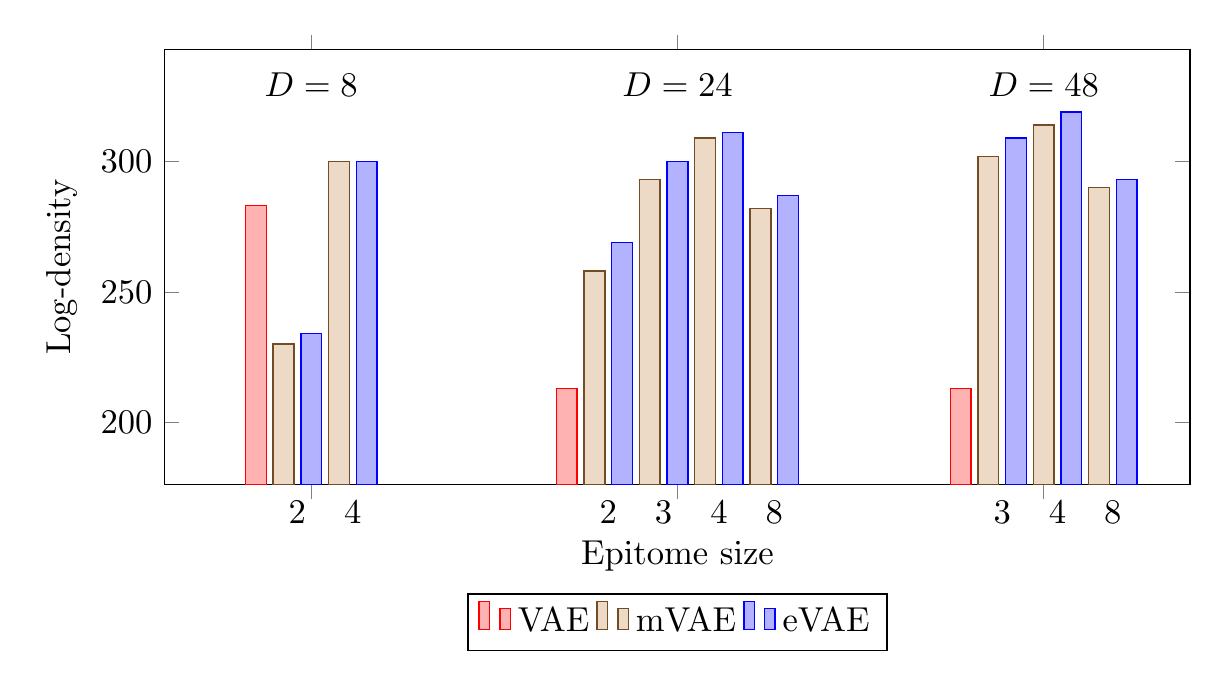}
    \caption{\small Epitome size vs. Parzen log-density in nats for different values of $D$ on MNIST. For each $D$, the optimal epitome size is significantly smaller than $D$.}
    \label{fig:epitome_size}
\end{figure}

We first qualitatively illustrate the ability of eVAE to overcome over-pruning and utilize latent capacity to model greater variability in data. Fig.~\ref{fig:mnist_generations} compares generation results for VAE and eVAE for different dimensions $D$ of latent variable $\bz$. With $D=2$, VAE generates realistic digits but suffers from lack of diversity. When $D$ is increased to $5$, the generation exhibits some greater variability but also begins to degrade in quality. As $D$ is further increased to $10$ and $20$, the degradation continues.  Contrast this with eVAE performance on generation: as the dimension $D$ of $\bz$ is increased while maintaining epitomes of size $K = 2$, eVAE is able to model greater variability in the data. Highlighted digits in the 20-d eVAE show multiple styles such as crossed versus un-crossed 7, and pointed, round, thick, and thin 4s. For both models, reconstruction improves with increasing $D$ as provided in Appendix Fig.~\ref{fig:mnist_reconstructions}.




\subsection{Choice of epitome size}

We next investigate how the choice of epitome size, $K$, affects generation performance.  We measure sample quality using the Parzen window estimator~\cite{rifai2012generative}. Fig.~\ref{fig:epitome_size} shows the Parzen log-density for different choices of epitome size on MNIST, with encoder and decoder consisting of a single deterministic layer of 500 units. Epitomes are non-overlapping, and the results are grouped by total dimension $D$ of the latent variable $\bz$. For comparison, we also show the log-density for VAE models with the same dimension $D$, and for mixture VAE (mVAE), an ablative version of eVAE where parameters are not shared. mVAE can also be seen as a mixture of independent VAEs trained in the same manner as eVAE. The number of deterministic units in each mVAE component is computed so that the total number of parameters is comparable to eVAE. 

As we increase $D$, the performance of VAE drops significantly, due to over-pruning. In fact, the number of active units for VAE are $8$, $22$ and $24$, corresponding to $D$ values of $8$, $24$ and $48$, respectively.   In contrast, eVAE performance increases as we increase $D$, with an epitome size $K$ that is significantly smaller than $D$.  This confirms the advantage of using eVAE to ensure good generation performance. Table~\ref{table:model_parameters} provides more comparisons. eVAE also performs comparably or better than mVAE at all epitome sizes. An explanation is that due to the parameter sharing in eVAE, each epitome benefits from general features learned across the training set.


\subsection{Increasing complexity of encoder and decoder}

Here we investigate the impact of encoder and decoder architectures with respect to over-pruning and generation performance. We vary model complexity through number of layers $L$ of deterministic hidden units, and number of hidden units $H$ in each deterministic layer. Table~\ref{table:model_parameters} shows the Parzen log-densities of VAE, mVAE and eVAE models trained on TFD with different latent dimension $D$ (See Appendix \S\ref{sec:mnist-complexity} for MNIST). All epitomes are non-overlapping and of size $K=5$. We observe that for VAE, increasing the number of hidden units $H$ (e.g. from 500 to 1000) for a fixed network depth $L$ has a negligible effect on the number of active units and performance. On the other hand, as the depth of the encoder and decoder $L$ is increased, the number of active units in VAE decreases though performance is still able to improve. This illustrates that increase in the complexity of the interactions through multiple layers counteract the perils of the over-pruning. However, this comes with the cost of substantial increase in the number of model parameters to be learned.

In contrast, for any given model configuration, eVAE is able to avoid the over-pruning effect in the number of active units and outperform VAE. Table~\ref{table:model_parameters} also shows results for mVAE, the ablative version of eVAE where parameters are not shared. The number of deterministic units per layer in each mVAE component is computed to have total number of parameters comparable to eVAE. These results are in line with the intuition that parameter sharing is helpful in more challenging settings when each epitome can also benefit from general features learned across the training set.

\begin{table}[ht]
\scriptsize
\begin{center}
\begin{tabular}{@{}ll|c c|c c}
\toprule
\multicolumn{2}{c}{} & \multicolumn{2}{c}{$H{=}500$} & \multicolumn{2}{c}{$H{=}1000$}\\
& & $L{=}2$ & $L{=}3$ & $L{=}2$ & $L{=}3$\\
\midrule
\multirow{2}{*}{$D{=}15$} & VAE & 2173(15) & 2180(15) & 2149(15) & 2116(15) \\
& mVAE & 2276(15) & 2314(15) & $\mathbf{2298}(15)$ & 2343(15) \\
& eVAE & $\mathbf{2298}(15)$ & $\mathbf{2353}(15)$ & 2278(15) & $\mathbf{2367}(15)$ \\
\midrule
\multirow{2}{*}{$D{=}25$} & VAE & 2067(25) & 2085(25) & 2037(25) & 2101(25) \\
& mVAE & 2287(25) & 2306(25) & $\mathbf{2332}(25)$ & 2351(25) \\
& eVAE & $\mathbf{2309}(25)$ & $\mathbf{2371}(25)$ & 2297(25) & $\mathbf{2371}(25)$ \\
\midrule
\multirow{2}{*}{$D{=}50$} & VAE & 1920(50) & 2062(29) & 1886(50) & 2066(30) \\
& mVAE & 2253(50) & 2327(50) & 2280(50) & 2358(50) \\
& eVAE & $\mathbf{2314}(50)$ & $\mathbf{2359}(50)$ & $\mathbf{2302}(50)$ & $\mathbf{2365}(50)$ \\
\end{tabular}
\end{center}
\caption{\small{Parzen log-densities in nats of VAE, mVAE and eVAE for increasing model capacity on TFD.  Across each row shows performance as the number of encoder and decoder layers $L$ increases for a fixed number of hidden units $H$ in each layer, and as $H$ increases. Number of active units are indicated in parentheses.}}
\label{table:model_parameters}
\end{table}

\subsection{Log-likelihood evaluation}
Table~\ref{table:likelihoods} shows importance weighted estimates as the mean of $\mathcal{L}_{5000}$ for VAE and eVAE on MNIST, with different dimensions $D$ of latent variable $z$. All models have 2 deterministic hidden layers of 200 units, and are trained in 8 stages with a learning rate of $0.001 \cdot 10^{\frac{-1}{7}}$ for $3^{i}$ epochs, for each stage $i = 0 ... 7$, following ~\citet{BGS16}. The VAE model has 20 active units at all $D$, so is not able to improve performance with increasing $D$. On the other hand, eVAE is able to leverage the additional latent capacity to improve on the log-likelihood. Note that these results can be improved through the tighter lower bound of IWAE~\citep{BGS16}, but this is an orthogonal consideration since epitomic training can also improve IWAE.

\begin{table}[ht]
\begin{center}
\begin{tabular}{l r r}
\toprule
D & VAE & eVAE\\
\midrule
$50$ & $86.76$ & $86.13$\\
$100$ & $86.91$ & $85.73$\\
$200$ & $87.09$ & $\mathbf{85.53}$\\
\end{tabular}
\end{center}
\caption{\small{Importance-weighted log-likelihood estimates as the mean of $\mathcal{L}_{5000}$ for VAE and eVAE on MNIST, with different dimensions $D$ of latent variable $z$. All models have 2 deterministic hidden layers of 200 units, and eVAE models use epitomes of size $K = 25$.  As $D$ increases, VAE is not able to take advantage of additional units to improve performance, while eVAE is.}}
\label{table:likelihoods}
\end{table}

\cut{
\begin{table}[ht]
\begin{center}
\begin{tabular}{l r r}
\toprule
Method & MNIST & TFD \\
\midrule
VAE & \textbf{91.8} / 66.03 / 25.78 & - \\
Mixture VAE & - & - \\
eVAE & \textbf{98.0} / 82.5 / 15.5 & - \\
\end{tabular}
\end{center}
\caption{\small{Lowerbound comparison for 50-d VAE and eVAE with single hidden layer of 500 units.}}
\label{table:lowerbound}
\end{table}
}

\subsection{Sample-based evaluation}
\begin{figure}
  \centering
  \begin{tabular}{@{}c@{\hskip 0.05in}c@{}}
  \subfloat[]{\includegraphics[width=0.48\linewidth]{{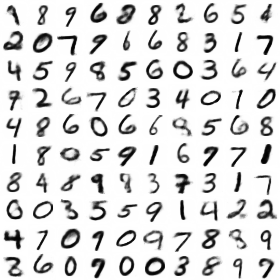}}} &
  \subfloat[]{\includegraphics[width=0.48\linewidth]{{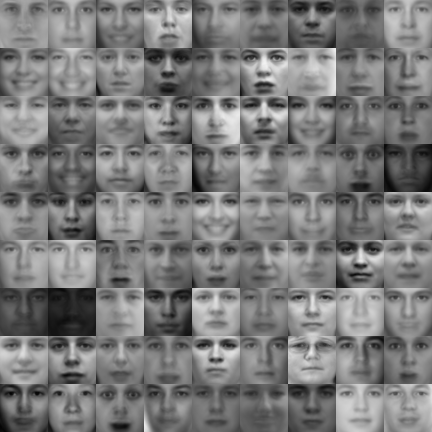}}}
  \end{tabular}
  \caption{\small{eVAE samples for: (a) MNIST, and (b) TFD.}}
  \label{fig:sample_generations}
\end{figure}

In Table~\ref{table:densities} we compare the generative performance of eVAE with other models through their samples. Encoders and decoders have $L=2$ layers of $H=1000$ deterministic units. $D=8$ for MNIST, and $D=15$ for TFD. VAE, mVAE, and eVAE refer to the best performing models over all architectures from Table~\ref{table:model_parameters}. For MNIST, the VAE model is $(L,H,D) = (3,500,8)$, mVAE is $(3,1000,24)$, and eVAE is $(3,500,48)$. For TFD, the VAE model is $(3,500,15)$, mVAE is $(3,1000,50)$, and eVAE is $(3,500,25)$. We observe that eVAE significantly improves over VAE and is competitive with several state-of-the-art models, notably Adversarial Autoencoders. Samples from eVAE on MNIST and TFD are shown in Fig.~\ref{fig:sample_generations}.
\cut{Table~\ref{table:lowerbound} reports the log likelihood of eVAE vs VAE on MNIST and TFD.}

\begin{table}[ht]
\begin{center}
\scriptsize
\begin{tabular}{l r r}
\toprule
Method & MNIST(10K) & TFD(10K) \\
\midrule
DBN~\citep{hinton2006fast} & $138 \pm 2$ & $1909 \pm 66$ \\
Deep CAE~\citep{bengio2013better} & $121 \pm 1$ & $2110 \pm 50$ \\
Deep GSN~\citep{thibodeau2014deep} & $214 \pm 1$ & $1890 \pm 29$ \\
GAN~\citep{goodfellow2014generative} & $225 \pm 2$ & $2057 \pm 26$ \\
GMMN + AE~\citep{li2015generative} & $282 \pm 2$ & $2204 \pm 20$ \\
Adversarial AE~\citep{MSJ5+} & $\mathbf{340 \pm 2}$ & $2252 \pm 16$ \\
\midrule
VAE & $325 \pm 2$ & $2180 \pm 20$ \\
mVAE & $\mathbf{338 \pm 2}$ & $2358 \pm 20$ \\
eVAE & $337 \pm 2$ & $\mathbf{2371 \pm 20}$ \\

\end{tabular}
\end{center}
\caption{\small{Parzen log-densities in nats on MNIST and TFD. VAE, mVAE, and eVAE refer to the best performing models over all architectures from Table~\ref{table:model_parameters}.}}
\label{table:densities}
\end{table}


\cut{
Key experiments:

\begin{itemize}
\item{Increased use of model capacity}
\item{Better generation results, as a function of dimensionality of z. comparison with VAE,}.
\item{Manifold traversal.} This is possible only if we get the model to train with overlapping epitome.
\item{Any notion of disentangled factors of variation?}
\item{Conditional generation}
\item{semi-supervised learning}
\end{itemize}

Datasets: MNIST, TFD

\subsection{Measuring latent  capacity}
In these experiments we study the effective dimensions. We visualize the number of active units that the models actually utilize. A unit $u$ is defined to be active if $A_u=\mathrm{Cov}_x(\mathbb{E}_{u \sim q(u|\mathbf{x})}[u]) > 0.02$, following~\cite{BGS16}. We also visualize reconstructions of test images using only subsets of the latent units based on how active they are.

\TODO{Compare with VAE, Dropout  VAE and our method}
}

\section{Related Work}
\label{sec:related} 


A number of applications use variational autoencoders as a building block.  In \cite{GDG+15}, a generative model for images is proposed in which the generator of the VAE is an attention-based recurrent model that is conditioned on the canvas drawn so far.  \cite{EHWTKH16} proposes a  VAE-based recurrent generative model that describes images as formed by sequentially choosing an object to draw and adding it to a canvas that is updated over time. In \cite{KWK+15}, VAEs are used for rendering 3D objects. Conditional variants of VAE are also used for attribute specific image generation \citep{YYS+15} and future frame synthesis \citep{XWB+16}. All these applications suffer from the problem of model over-pruning and hence have adopted strategies that takes away the clean mathematical formulation of VAE. We have discussed these in \S~\ref{sec:vae_prune}. A complementary approach to the problem of model pruning in VAE was proposed in  \cite{BGS16}; the idea is to improve the variational bound by using  multiple weighted posterior samples. Epitomic VAE provides improved latent capacity even when only a single sample is drawn from the posterior.
 
Methods to increase the flexibility of posterior inference are proposed in \citep{SKW+15,RM+16, KSW+16}. In \cite{RM+16}, posterior approximation is constructed by transforming a simple initial density into a complex one with a sequence of invertible transformations. \cite{KSW+16} augments the flexibility of the posterior through autoregression over projections of stochastic latent variables. However, the problem of over-pruning still persists: for instance, \cite{KSW+16}  enforces a minimum information constraint to ensure all units are used. 

Related is research in unsupervised sparse overcomplete representations, especially with group sparsity constraints {\it c.f.} \citep{Gregor11,JM+11}. In the epitomic VAE, we have similar motivations that enable learning better generative models of data. 





\section{Conclusion}
\label{sec:conclusion}

This paper introduces Epitomic VAE, an extension of variational autoencoders, to address the problem of model over-pruning, which has limited the generation capability of VAEs in high-dimensional spaces. Based on the intuition that subconcepts can be modeled with fewer dimensions than the full latent space, epitomic VAE models the latent space as multiple shared subspaces that have learned specializations. We show how this model addresses the model over-pruning problem in a principled manner, and present qualitative and quantitative analysis of how eVAE enables increased utilization of the model capacity to model greater data variability. We believe that modeling the latent space as multiple structured subspaces is a promising direction of work, and allows for increased effective capacity that has potential to be combined with methods for increasing the flexibility of posterior inference.

\section{Acknowledgments}
We thank Marc'Aurelio Ranzato, Joost van Amersfoort and Ross Girshick for helpful discussions. We also borrowed the term `epitome' from an earlier work of \cite{Jojic:2003}.

\bibliography{epitomicVAE}
\bibliographystyle{icml2017}

\clearpage
\section{Appendix}
\label{sec:appendix}

\subsection{Effect of KL weight $\lambda$ on reconstruction}
\label{sec:appendix-lambda}

We visualize VAE reconstructions as the KL term weight $\lambda$ is tuned down to keep latent units active. The top half of each figure are the original digits, and the bottom half are the corresponding reconstructions. While reconstruction performance is good, generation is poor (Fig.~\ref{fig:kld_lambda}). This illustrates that VAE learns to model well only regions of the posterior manifold near training samples, instead of generalizing to model well the full posterior manifold.

\begin{figure}[h]
  \centering
  \begin{tabular}{@{}c@{\hskip 0.2in}c@{\hskip 0.2in}c@{}}
   \subfloat[$\lambda=1.0$]{\includegraphics[width=0.3\linewidth]{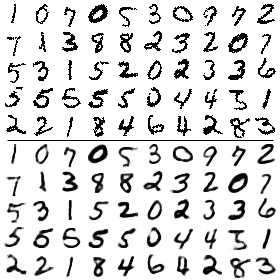}} &
  \subfloat[$\lambda=0.5$]{\includegraphics[width=0.3\linewidth]{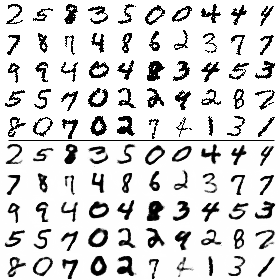}} &
  \subfloat[$\lambda=0.2$]{\includegraphics[width=0.3\linewidth]{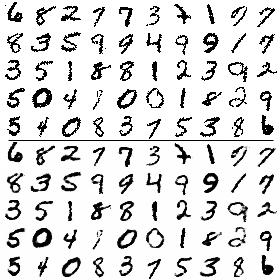}}
  \end{tabular}
  \caption{\small{Reconstructions for a 50-d VAE with KL weight $\lambda = 1$, $0.5$, and $0.2$. The top half of each figure are the original digits, and the bottom half are the corresponding reconstructions.}}
  \label{fig:reconstruction_lambda}
\end{figure}


\subsection{Effect of increasing latent dimension on reconstruction}

\begin{figure*}
	\centering
\includegraphics[width=1.0\linewidth]{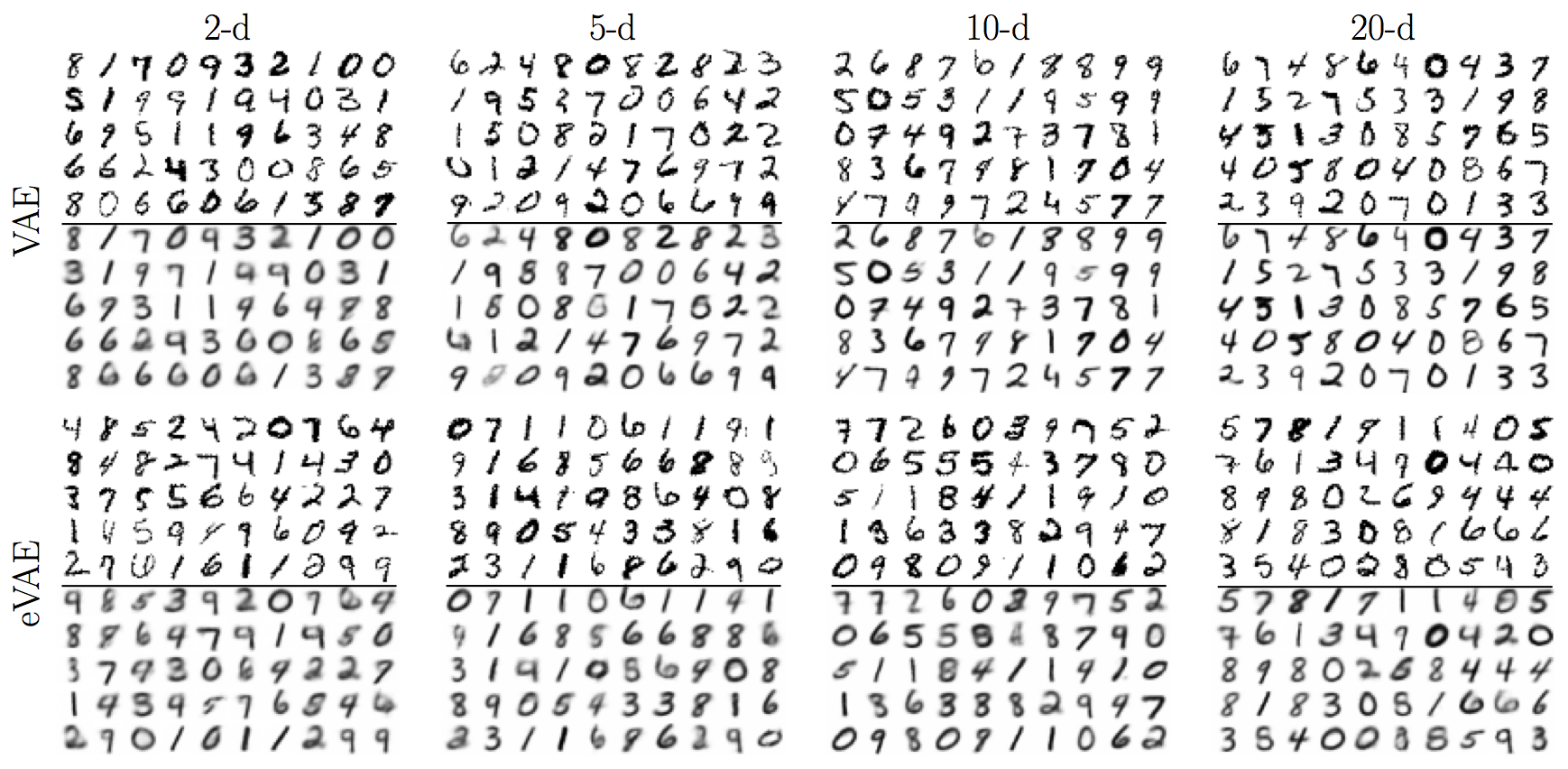}
    \caption{\small Reconstructions from VAE and eVAE models for different dimensions of latent variable $\bz$. Across each row are 2-d, 5-d, 10-d, and 20-d models. The top half of each figure are the original digits, and the bottom half are the corresponding reconstructions. The eVAE models multiple shared subspaces by maintaining 2-d (overlapping) epitomes as the latent dimension is increased. In contrast to VAE, eVAE achieves both good reconstruction and generation.}
  	\label{fig:mnist_reconstructions}
\end{figure*}

In \S~\ref{sec:exp-overpruning} in the main paper, Fig.~\ref{fig:mnist_generations} shows the effect of increasing latent dimension on generation for VAE and eVAE models. Here we show the effect of the same factor on reconstruction quality for the models (Fig.~\ref{fig:mnist_reconstructions}). The top half of each figure are the original digits, and the bottom half are the corresponding reconstructions. As the dimension of the latent variable $\bz$ increases from 2-d to 20-d, VAE reconstruction becomes very sharp (the best model), but generation degrades (Fig.~\ref{fig:mnist_generations}). On the other hand, eVAE is able to achieve both good reconstruction and generation.

\subsection{MNIST encoder and decoder complexity}
\label{sec:mnist-complexity}

Table~\ref{table:model_parameters_mnist} shows the effect of increasing complexity in the encoder and decoder complexity on MNIST. We vary model complexity through number of layers $L$ of deterministic hidden units, and number of hidden units $H$ in each deterministic layer. Parzen log-densities are provided for VAE, mVAE and eVAE models trained on MNIST with different latent dimension $D$. The effect of model complexity on active units and performance aligns with that for TFD in the main paper.

\begin{table*}
\scriptsize
\begin{center}
\begin{tabular}{l l | c c c | c c c}
\toprule
\multicolumn{2}{c}{} & \multicolumn{3}{c}{$H=500$} & \multicolumn{3}{c}{$H=1000$}\\
& & $L=1$ & $L=2$ & $L=3$ & $L=1$ & $L=2$ & $L=3$\\
\midrule
\multirow{3}{*}{$D=8$} & VAE & $283 (8)$ & $292 (8)$ & 325(8) & 283(8) & 290(8) & 322(6) \\
& mVAE & $\mathbf{300}(8)$ & 328(8) & $\mathbf{337}(8)$ & 309(8) & $\mathbf{333}(8)$ & $\mathbf{335}(8)$ \\
& eVAE & $\mathbf{300} (8)$ & $\mathbf{330}(8)$ & $\mathbf{337}(8)$ & $\mathbf{312}(8)$ & 331(8) & 334(8) \\
\midrule
\multirow{3}{*}{$D=24$} & VAE & 213(22) & 273(11) & 305(8) & 219(24) & 270(12) & 311(7) \\
& mVAE & 309(24) & 330(24) & $\mathbf{336}(24)$ & 313(24) & $\mathbf{333}(24)$ & $\mathbf{338}(24)$ \\
& eVAE & $\mathbf{311}(24)$ & $\mathbf{331}(24)$ & $\mathbf{336}(24)$ & $\mathbf{317}(24)$ & $\mathbf{332}(24)$ & $\mathbf{336}(24)$ \\
\midrule
\multirow{3}{*}{$D=48$} & VAE & 213(24) & 267(13) & 308(8) & 224(24) & 273(12) & 309(8) \\
& mVAE & 314(48) & $\mathbf{334}(48)$ & 336(48) & 315(48) & 333(48) & $\mathbf{337}(48)$ \\
& eVAE & $\mathbf{319}(48)$ & $\mathbf{334}(48)$ & $\mathbf{337}(48)$ & $\mathbf{321}(48)$ & $\mathbf{334}(48)$ & $\mathbf{332}(48)$ \\
\end{tabular}
\end{center}
\caption{\small{Parzen log-densities in nats of VAE, mVAE and eVAE for increasing model parameters, trained on MNIST with different dimensions $D$ of latent variable $\bz$.  For mVAE and eVAE models, the maximum over epitomes of size $K=3$ and $K=4$ is used. All epitomes are non-overlapping. Across each row shows performance as the number of encoder and decoder layers $L$ increases for a fixed number of hidden units $H$ in each layer, and as $H$ increases. Number of active units are indicated in parentheses.}}
\label{table:model_parameters_mnist}
\end{table*}

\end{document}